\documentclass[a4paper,twoside]{article}

\usepackage{epsfig}
\usepackage{subcaption}
\usepackage{calc}
\usepackage{amssymb}
\usepackage{amstext}
\usepackage{amsmath}
\usepackage{amsthm}
\usepackage{multicol}
\usepackage{pslatex}
\usepackage{apalike}
\usepackage{algorithm2e}
\usepackage[bottom]{footmisc}

\usepackage{booktabs} % To thicken table lines
\usepackage{graphicx}
\usepackage{hyperref}
\usepackage{SCITEPRESS}     

\begin{document}

\title{LostPaw: Finding Lost Pets using a Contrastive Learning-based Transformer with Visual Input}

%%%%%%%%%%%%%%%%%%%%%%%%%%%%%%%%%%%%%%%%%%%%%%%%%%%%%%%%%%%%%%%%%%%%%%%%
%%%%%%%%%%%%%%%%%%%%%%%%%%%%%%%%%%%%%%%%%%%%%%%%%%%%%%%%%%%%%%%%%%%%%%%%
%%%% Authors removed for double-blind
\author{\authorname{Andrei Voinea, Robin Kock and Maruf A. Dhali\orcidAuthor{0000-0002-7548-3858}}
\affiliation{Department of Artificial Intelligence, Bernoulli Institute, University of Groningen, 9747 AG Groningen, The Netherlands}
\email{\{c.a.voinea@student.rug.nl,r.kock@student.rug.nl, m.a.dhali@rug.nl}
}

%%%%%%%%%%%%%%%%%%%%%%%%%%%%%%%%%%%%%%%%%%%%%%%%%%%%%%%%%%%%%%%%%%%%%%%%
%%%%%%%%%%%%%%%%%%%%%%%%%%%%%%%%%%%%%%%%%%%%%%%%%%%%%%%%%%%%%%%%%%%%%%%%


\abstract{Losing pets can be highly distressing for pet owners, and finding a lost pet is often challenging and time-consuming. An artificial intelligence-based application can significantly improve the speed and accuracy of finding lost pets. To facilitate such an application, this study introduces a contrastive neural network model capable of accurately distinguishing between images of pets. The model was trained on a large dataset of dog images and evaluated through 3-fold cross-validation. Following 350 epochs of training, the model achieved a test accuracy of 90\%. Furthermore, overfitting was avoided, as the test accuracy closely matched the training accuracy. Our findings suggest that contrastive neural network models hold promise as a tool for locating lost pets. This paper presents the foundational framework for a potential web application designed to assist users in locating their missing pets. The application will allow users to upload images of their lost pets and provide notifications when matching images are identified within its image database. This functionality aims to enhance the efficiency and accuracy with which pet owners can search for and reunite with their beloved animals.}

\onecolumn \maketitle \normalsize \setcounter{footnote}{0} \vfill

%%%%%%%%%%%%%%%%%%%%%%%INTRODUCTION%%%%%%%%%%%%%%%%%%%%%%%%%%%%%

\section{\uppercase{Introduction}}
Experiencing the loss of a pet can be deeply traumatic for owners, who frequently have difficulty locating them even after utilizing flyers, online searches, and private investigators. Since pets can roam far from home, conventional search strategies often prove ineffective. Without unified communication channels, assistance can be limited. Artificial intelligence can help identify lost pets through image analysis, though comparing images can still be challenging for volunteers.

In recent years, contrastive learning has emerged as a promising solution to the problem of differentiating between two or more input data classes in computer vision \cite{chenSimple2020}. This approach involves training a machine learning model to identify subtle differences between images by comparing pairs of data samples. This technique has demonstrated notable efficacy in various visual recognition tasks, such as image classification, where models are trained to differentiate between objects or scenes based on visual features.

Contrastive learning methods efficiently learn high-dimensional data representations by comparing classes, enabling them to handle transformations like rotation and scaling. For instance, a model trained on pet images can help identify missing animals, aiding owners and shelters. Analyzing image pairs allows a contrastive neural network to distinguish between breeds and individuals, improving lost pet searches. This paper will explore the technical aspects and applications of this approach to develop a complete web-based solution.

%%%%%%%%%%%%%%%%%%%%%%%BACKGROUND%%%%%%%%%%%%%%%%%%%%%%%%%%%%%

\section{\uppercase{Related Works}}
Several components are required to create a contrastive learning model capable of differentiating between images of pets. A fundamental part of such a model is a neural network architecture that can learn a robust and effective data representation. In this study, we employed the Vision Transformer model as the foundation of our contrastive learning model. In addition, we used the Detection Transformer model to extract the pets from the images and the AutoAugment feature to augment the images. Finally, to optimize the model, we utilize a contrastive loss function, which allows the model to learn the underlying structure of the data by contrasting similar and dissimilar examples. In the following sections, we provide a more in-depth description of these technologies and their implementation in our contrastive learning transformer model.

\subsection{Transformer Models}
Transformer models are a type of neural network architecture widely successful in various natural language processing tasks and have achieved state-of-the-art results on a large selection of benchmarks \cite{vaswaniAttention2017a}. Self-attention mechanisms enable the model to focus on different parts of input data at various times. This allows it to capture long-range dependencies, especially for tasks like language translation, where a word's meaning depends on context. Additionally, transformers utilize multi-headed attention, allowing simultaneous focus on multiple data parts. This enhances processing capabilities and overall performance on numerous tasks, including natural language processing, image classification, and object detection.

\subsection{Detection Transformer}
The Detection Transformer (DETR) is an end-to-end object detector that uses a Transformer encoder-decoder architecture \cite{carionEndtoEnd2020}. It features a convolutional neural network (CNN) backbone that extracts image features, which are flattened and combined with positional encoding before being processed by the Transformer encoder to generate feature maps representing objects. The output from the encoder is fed into a Transformer decoder that uses learned positional embeddings called \textit{object queries}. The decoder creates embeddings for each query, which are then classified as detections or "no object" via a shared feedforward network. When detection occurs, the model provides the object's class (e.g., cat or dog) and the bounding box indicating its location in the image.

\subsection{Vision Transformer}
The Vision Transformer (ViT) is a neural network architecture for image classification that processes raw pixel values instead of using convolutional layers \cite{dosovitskiyImage2021}. It consists of transformer blocks with self-attention mechanisms that analyze 16×16 pixel patches from images, allowing the model to determine the importance of different patches based on their relationships. The patches are embedded into a high-dimensional space and processed through twelve transformer blocks, which output a new sequence of patches. The output is then passed through a linear layer and a softmax function to produce final class probabilities. ViT is trained with supervised learning, using ground truth labels to calculate cross-entropy loss for weight updates. This enables models trained for classification tasks to be fine-tuned for various applications, including image pair comparisons.

\subsection{AutoAugment}
AutoAugment is a method that automates data augmentation, which involves applying transformations to images in a dataset to increase its size and enhance the robustness of machine learning models \cite{cubukAutoAugment2019}. It frames the task of finding the optimal augmentation policy as a discrete search problem, utilizing a recurrent neural network as a controller to sample policies that dictate which image operations to apply, their probabilities, and their intensity.

The algorithm is trained using policy gradient methods, allowing it to adjust based on the validation accuracy of a neural network trained with a fixed architecture. AutoAugment offers various pre-configured policies with transformation functions, including shearing, translation, rotation, and adjustments in contrast, brightness, and sharpness. This approach effectively enhances data augmentation and improves the performance of models, especially in image classification tasks.

\subsection{Contrastive Loss}
Contrastive loss is a loss function widely used in machine learning for unsupervised learning \cite{hadsellDimensionality2006}. It aims to learn data representations highlighting class relationships and differences between similar and dissimilar examples. This loss function is often applied with Siamese networks, which consist of two or more identical subnetworks that process different inputs \cite{kochSiamese2015}. Trained with the exact weights, these subnetworks learn a shared data representation. The contrastive loss is then computed from this representation, guiding updates to the network's weights.

As proposed by \cite{hadsellDimensionality2006}, the contrastive loss function, shown in Equation~\ref{eq:loss}, uses $d$ to calculate the Euclidean distance between two pet vectors and $m$ as a margin that controls sensitivity in classifying images as similar. The function minimizes the distance between feature vectors of the same pet while maximizing the distance between those of different pets. Furthermore, the loss function ensures that the distance between the feature vectors of dissimilar examples exceeds the margin given by the hyperparameter $m$.
\begin{multline}    
\mathcal{L}(X_1, X_2) = \\
\frac{1}{2} \hspace{1mm}
\text{\large E}\left[\begin{cases}
    \max \{ m - d(X_1, X_2), 0 \}^2 & \text{different pet}\\
    d(X_1, X_2)^2 & \text{same pet}
\end{cases}\right] \\
\label{eq:loss}
\end{multline}

%%%%%%%%%%%%%%%%%%%%%%%METHODS%%%%%%%%%%%%%%%%%%%%%%%%%%%%%

\section{\uppercase{Methodology}}
In this study, we develop a contrastive neural network model to distinguish between pet images. Trained on a large dataset of various dog breeds, the model is evaluated on a separate test set. Implemented using the PyTorch framework \cite{paszke2017automatic}, it utilizes supervised learning techniques. The following subsection will cover the dataset, model architecture, training procedure, and evaluation metrics used to assess performance.

\subsection{Dataset}
\begin{figure}[!ht]
    \centering
    \includegraphics[width=\linewidth]{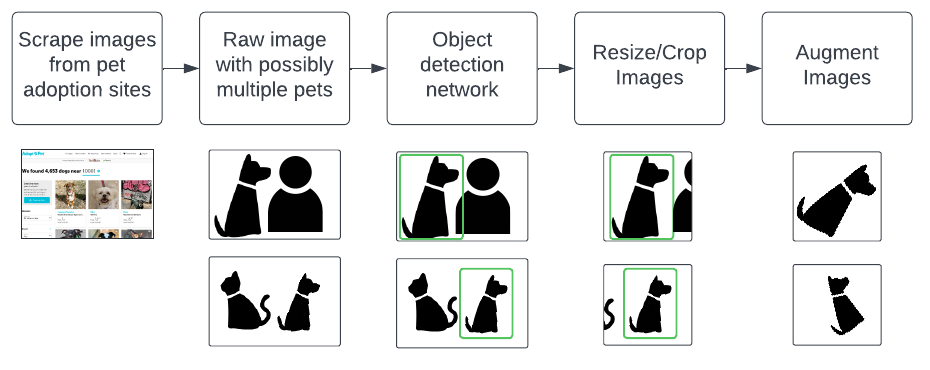}
    \caption{Data collection process. The top nodes represent the individual steps that are taken for each image. The diagrams at the bottom show possible configurations of each step.}
    \label{fig:data_pipeline}
\end{figure}

To create the dataset used in this study, we obtain images of pets from adoption websites such as AdoptAPet \cite{adoptapet}. Each image is fed through the DETR model, and the resulting bounding boxes of pets are used to crop them from the image. We focus only on images of dogs, resulting in $31,860$ pets being stored with an average of $2.47$ images per pet ($78,702$ total images). The cropped images are then resized to fit a square of $384\times384$ pixels, and if the image is not wide or tall enough, the missing area is filled with black. Each image is then augmented twice using a pre-trained AutoAugment model, which follows the policies CIFAR10, ImageNet, and SVHN. A test set is created by setting aside images extracted from $3595$ pets, totaling $8854$ images. The augmented dataset contained $236,106$ train images and $26,562$ test images, which are used to train and evaluate the contrastive neural network model developed in this study. For a schematic view of the data pipeline, see Figure~\ref{fig:data_pipeline}.

To enable the use of our data for contrastive learning, we need to further combine the images into pairs, forming a pairwise dataset. Each pair is labeled as either \textit{different} or \textit{same} and contains two images of size $384\times384$. The pairwise dataset is compiled using a random number generator to select labels and images, and a seed value is used to ensure that the dataset is reproducible between runs.

To sample a pair during the dataset generation, we first choose whether the label is \textit{same} or \textit{different} based on a similarity probability. A value of 50\% is chosen for this probability to create an approximately equal number of the same and different pairs. By selecting pairs to be the same or different with equal probability, we ensure that the contrastive ViT model is exposed to a balanced set of training labels. This can help prevent the model from becoming biased towards one type of example or the other and improve the model's generalization performance on the held-out test set.

\begin{figure}[hbt!]
    \centering
    \includegraphics[width=\linewidth]{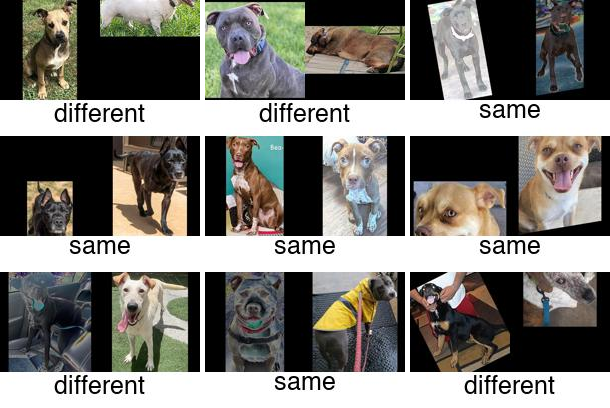}
    \caption{Example data pairs with labels underneath. Some of the images have been augmented.}
    \label{fig:data}
\end{figure}

Following this, the images for the pair are selected from the cropped image set described earlier. If the label is \textit{different}, we randomly select two different pets from the dataset (using the same seed described earlier) and then select an image for each pet. If the label is \textit{same}, a random pet is chosen, and the pair of images comprises two different images of the same pet. Furthermore, as each image in the dataset is augmented twice, we ensure that we never choose two augmentations of the same image. The process of selecting pairs is repeated to generate an extensive set of training pairs for the contrastive ViT model, an example of which can be observed in Figure \ref{fig:data}.

As the dataset is dynamically generated during training, validating a model requires special attention due to its stochastic nature. In this study, we employ $k$-fold cross-validation with a pairwise dataset, assigning each pair to one of $k$ distinct folds. This ensures that during testing, each fold is used only once while the remaining folds are for training, preventing exposure to the same pairs. However, since pairs are formed from images, there is a possibility that the same images may appear in both training and cross-validation. Due to the random selection process, it’s possible for an image to be compared to one it has encountered before. Nevertheless, the large size of the dataset makes duplicate pairs unlikely. To mitigate this, we present additional results from a held-out test set featuring entirely novel pets.

\begin{figure}[!ht]
    \centering
    \includegraphics[width=\linewidth]{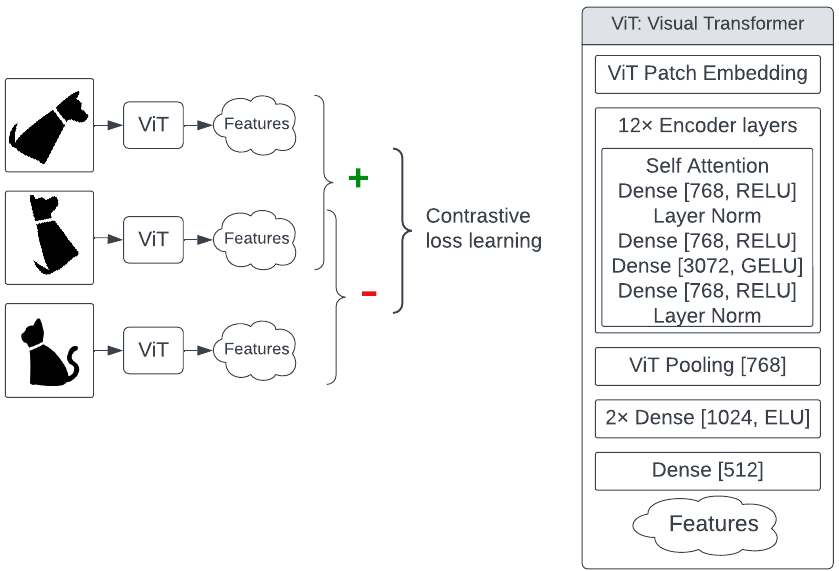}
    \caption{Architecture of the Contrastive Vision Transformer model.}
    \label{fig:model_architecture}
\end{figure}

\subsection{Model Architecture}
To develop the contrastive ViT model, we use ViT as the backbone of the model \cite{wu2020visual}. The ViT output is flattened and followed by three fully connected layers to achieve the desired latent vector size. The model features two hidden layers with twice the number of neurons as the latent space and a final output layer matching the latent space size. These layers transform the ViT backbone's output into a compact representation of the image's structure. We use the Exponential Linear Unit (ELU) activation function between the layers to propagate negative values for the contrastive loss function. For a graphical overview of the model architecture, see Figure~\ref{fig:model_architecture}.

During training, we fine-tune only the last three layers of the model while keeping the backbone parameters frozen. This approach, while limiting the model’s ability to learn new lower-level features, results in a more stable training process and reduces the risk of overfitting. An ablation study where all parameters are updated revealed a significant drop in performance, which is discussed in the Results section. We select the hyperparameters for the contrastive ViT model through parameter sweeps. By training with various values and evaluating the model on a validation set, we identify the optimal hyperparameters. This process helps us fine-tune the model to better differentiate between pictures of pets. Table \ref{tab:hyper} lists these hyperparameters.

\begin{table}[hbt!]
    \centering
    \caption{Hyperparameters for the contrastive ViT model.}
\resizebox{0.65\linewidth}{!}{
    \begin{tabular}{ll}
        \toprule
        Name & Value \\ 
        \midrule
        Epochs & 350 \\
        Latent Space Size & 512 \\
        Batch Size & 8 \\
        Batch Count per Epoch & 128 \\
        Test Batch Size & 8 \\
        Test Batch Count & 128 \\
        Optimizer & AdamW \\
        Learning Rate & 5.0e-5 \\
        Weight Decay & 2.0e-4 \\
        Contrastive Margin & 1.66 \\
        \bottomrule
    \end{tabular}}
    \label{tab:hyper}
\end{table}

\subsection{Evaluation Metrics}
We use $k$-fold cross-validation to evaluate the performance of the contrastive learning model. We use $k=3$ for our cross-validation, which resulted in 3 different models being trained and evaluated. We trained the model for a fixed number of epochs for each fold and used the validation set to tune the model's hyperparameters. Once the model was trained, we evaluated it on the test set and recorded its performance in terms of accuracy, the type I and type II errors, and the $F_1$ score. The type I error represents the proportion of false positives, while the type II error represents the probability of false negatives. Using these values, we calculate the precision and recall of our model, which are used to obtain the $F_1$ score value (see Equation \ref{eq:f1-score}). The precision and recall values represent the performance of the classification model on the given dataset.

\begin{equation}
\label{eq:f1-score}
F_1 = 2\cdot\frac{\text{Precision}\cdot\text{Recall}}{\text{Precision}+\text{Recall}}
\end{equation}

%%%%%%%%%%%%%%%%%%%%%%%RESULTS%%%%%%%%%%%%%%%%%%%%%%%%%%%%%

\section{\uppercase{Results}}
\label{sec:results}
\begin{figure}[hbt!]
    \centering
    \begin{subfigure}{0.38\textwidth}
        \includegraphics[width=\linewidth]{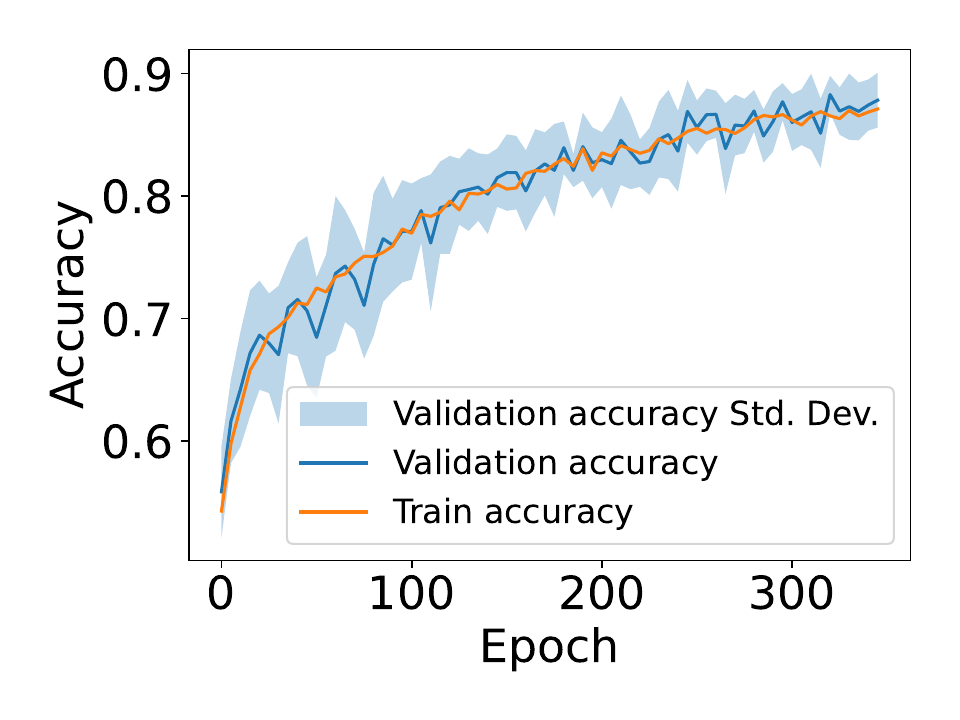}
        \caption{Accuracy}
        \label{fig:accuracy}
    \end{subfigure}
    \hfill
    \begin{subfigure}{0.4\textwidth}
        \includegraphics[width=\linewidth]{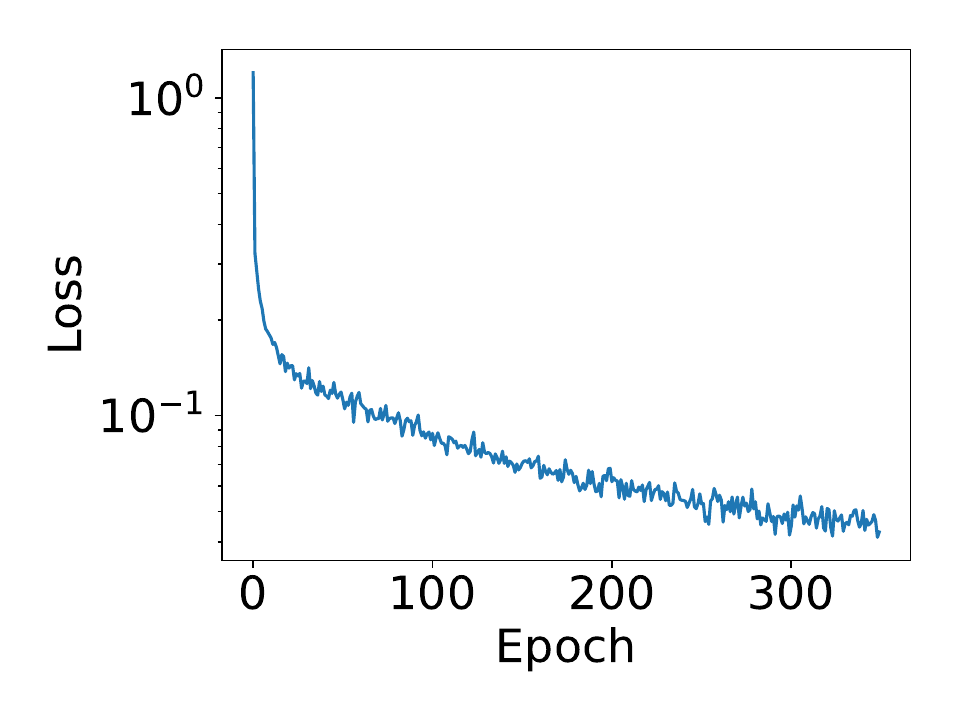}
        \caption{Loss}
        \label{fig:loss}
    \end{subfigure}
    \caption{Mean train accuracy and loss of the contrastive ViT model, averaged over three model runs. The data for accuracy was smoothed by averaging the values every five epochs.}
    \label{fig:acc_loss}
\end{figure}

Throughout 350 epochs, an average $F_1$ score of 88.8\% on the cross-validation set was achieved. In addition, the model was trained on a large dataset and did not appear to be overfitting, as the validation accuracy closely followed the training accuracy (see Figure~\ref{fig:accuracy}).

In addition to the accuracy results, we further examined the loss values of the model during training (see Figure \ref{fig:loss}). We observed that the loss value steadily dropped from a starting value of approximately 1.16 to a final value of approximately 0.04. This trend generally indicates the model is steadily learning a better representation of the data throughout training. Furthermore, the low loss value suggests that the model was able to learn an adequate representation of the data, which could potentially allow the model to make accurate decisions on unseen samples.

\begin{figure}[hbt!]
    \centering
    \includegraphics[width=0.8\linewidth]{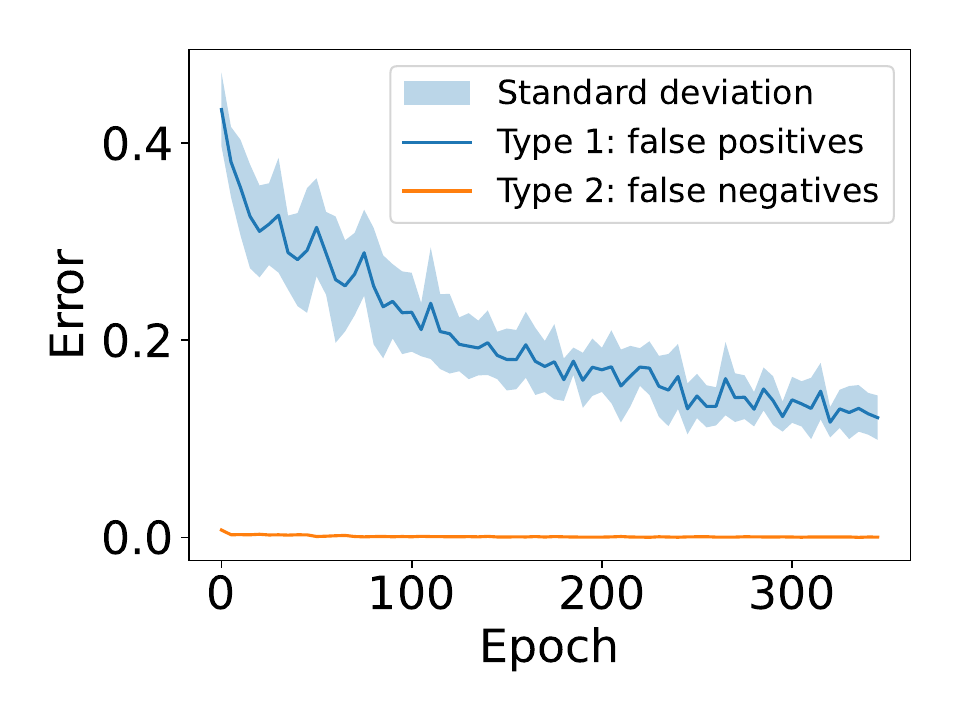}
    \caption{Type I and II errors of the model on the test set at every epoch. The data for the errors were smoothed by averaging the values every five epochs.}
    \label{fig:error-rates}
\end{figure}

When examining the errors of the model (see Figure~\ref{fig:error-rates}), we observed that the model initially classified every pair of pet images as the same pet. However, over the course of training, the model learned to differentiate between different pets, and the type I error decreased. Furthermore, the type II error was very close to zero for most of the training period. These results suggest that the model could learn a robust and relatively effective representation of the data, which could distinguish between different pets.

\begin{table}[ht!]
	\centering
  	\caption{Accuracy \& errors of the model for various sets.}
    \resizebox{\linewidth}{!}{
	\begin{tabular}{p{19.3mm}p{12mm}p{10mm}p{10mm}p{13mm}}
		\toprule
		&\multicolumn{4}{c}{Dataset}                   \\
		\cmidrule(r){2-5}
		Metric & Training & Cross-val. & Held-out & Held-out std.\\
		\midrule
		Accuracy      & 0.8687 & 0.8737 & 0.9028 & 0.0036 \\
		Type I error  & 0.1309 & 0.1261 & 0.0966 & 0.0036 \\
		Type II error & 0.0004 & 0.0002 & 0.0006 & 0.0003 \\
		$F_1$ score   & 0.8838 & 0.8880 & 0.9108 & 0.0041 \\
		\bottomrule
	\end{tabular}}
	\label{tab:results}
\end{table}

When examining the outcomes of the models on the held-out test set, as illustrated in Table \ref{tab:results}, we noted that the average $F_1$ score was 91.1\% (SD=0.41\%). Similarly, the mean type I error was 9.7\%, and the type II error was 0.06\%. These outcomes indicate an improvement over the metric values recorded on the train and validation sets. This may be attributed to various reasons, which are discussed in detail in the subsequent section. However, the results still suggest that the model has effectively generalized to new data.

Furthermore, during the ablation study, we observed that a fully-trained contrastive ViT model achieved a cross-validation $F_1$ score of 80.0\% and a held-out test set $F_1$ score of 78.6\%. This provides strong support for our decision to set the layers of the backbone model as fixed. In particular, the model appears to overfit the data more than it does with frozen layers since the held-out test set performance is inferior to the validation performance. For more results of the ablation study, see Figure \ref{fig:abl_results} in Appendix \ref{sec:abl_appendix}.

%%%%%%%%%%%%%%%%%%%%%%%CONCLUSION%%%%%%%%%%%%%%%%%%%%%%%%%%%%%

\section{\uppercase{Conclusions}}
This study involved the development of a contrastive neural network model to distinguish between images of dogs, with its performance evaluated on a held-out test set. Results from a three-fold cross-validation indicated that the model can accurately differentiate between pet images and generalize well to unseen data. The following section provides a detailed discussion of the evaluation results and the implications of these findings for applying artificial intelligence in searching for lost pets.

One concern is the relatively high incidence of false positives. While this may initially appear to be a limitation, it could be beneficial in locating lost pets, particularly when a few pets are reported missing in a given area, allowing for easy dismissal of incorrect identifications. Moreover, the use of the AutoAugment feature, which sometimes alters the colors of pet images, may influence the accuracy of the model. However, this variation could enhance generalization by allowing the model to learn more robust features, improving its performance on real-world data with varying color and lighting conditions.

A potential issue has been identified in the model's accuracy metrics: both the cross-validation accuracy and the held-out test set accuracy are higher than the training accuracy, which needs further investigation. The higher cross-validation accuracy may result from random fluctuations, as the two accuracies often intersect during training, as shown in Figure \ref{fig:accuracy}. The improvement of over 1\% in the held-out test set's performance compared to the cross-validation set is unclear and could be due to differences in data distribution or the smaller sample size in the held-out set. However, measures have been taken to eliminate any systematic errors that might affect the observed performance gains.

Future research could explore expanding the network to include various types of pets. This might involve using the DETR to identify the specific pet in an image, such as a cat or dog, and then passing it to a fine-tuned model that specializes in comparing pets within each category. This method would combine the strengths of both DETR and ViT models, resulting in a more robust system through enhanced contrastive data. Additionally, while this study focused on dog images, the described contrastive learning approach can be applied to other datasets. Training on diverse images enables the model to differentiate between various classes, with potential applications in medical image classification, wildlife species identification, and handwriting comparison. Overall, the study suggests that contrastive learning can significantly improve image classification accuracy.

%%%%%%%%%%%%%%%%%%%%%%%bibliography%%%%%%%%%%%%%%%%%%%%%%%%%%%%%
\bibliographystyle{apalike}
{\small
% \bibliography{bib}}

\begin{thebibliography}{}

\bibitem[Carion et~al., 2020]{carionEndtoEnd2020}
Carion, N., Massa, F., Synnaeve, G., Usunier, N., Kirillov, A., and Zagoruyko, S. (2020).
\newblock End-to-{{End Object Detection}} with {{Transformers}}.
\newblock In Vedaldi, A., Bischof, H., Brox, T., and Frahm, J.-M., editors, {\em Computer {{Vision}} \textendash{} {{ECCV}} 2020}, Lecture {{Notes}} in {{Computer Science}}, pages 213--229, {Cham}. {Springer International Publishing}.

\bibitem[Chen et~al., 2020]{chenSimple2020}
Chen, T., Kornblith, S., Norouzi, M., and Hinton, G. (2020).
\newblock A simple framework for contrastive learning of visual representations.
\newblock In III, H.~D. and Singh, A., editors, {\em Proceedings of the 37th International Conference on Machine Learning}, volume 119 of {\em Proceedings of Machine Learning Research}, pages 1597--1607. PMLR.

\bibitem[Cubuk et~al., 2019]{cubukAutoAugment2019}
Cubuk, E.~D., Zoph, B., Mane, D., Vasudevan, V., and Le, Q.~V. (2019).
\newblock {AutoAugment}: {Learning Augmentation Strategies from Data}.
\newblock In {\em 2019 IEEE/CVF Conference on Computer Vision and Pattern Recognition (CVPR)}, pages 113--123. IEEE Computer Society.

\bibitem[Dosovitskiy et~al., 2021]{dosovitskiyImage2021}
Dosovitskiy, A., Beyer, L., Kolesnikov, A., Weissenborn, D., Zhai, X., Unterthiner, T., Dehghani, M., Minderer, M., Heigold, G., Gelly, S., Uszkoreit, J., and Houlsby, N. (2021).
\newblock An {{Image}} is {{Worth}} 16x16 {{Words}}: {{Transformers}} for {{Image Recognition}} at {{Scale}}.

\bibitem[Hadsell et~al., 2006]{hadsellDimensionality2006}
Hadsell, R., Chopra, S., and LeCun, Y. (2006).
\newblock Dimensionality {{Reduction}} by {{Learning}} an {{Invariant Mapping}}.
\newblock In {\em 2006 {{IEEE Computer Society Conference}} on {{Computer Vision}} and {{Pattern Recognition}} ({{CVPR}}'06)}, volume~2, pages 1735--1742.

\bibitem[Inc and Affiliates, 2023]{adoptapet}
Inc, K.~P. and Affiliates (2023).
\newblock {AdoptAPet: Search for local pets in need of a home}.
\newblock https://www.adoptapet.com/, last visited: 28 April 2023.

\bibitem[Koch et~al., 2015]{kochSiamese2015}
Koch, G., Zemel, R., Salakhutdinov, R., et~al. (2015).
\newblock Siamese neural networks for one-shot image recognition.
\newblock In {\em ICML deep learning workshop}, volume~2. Lille.

\bibitem[Paszke et~al., 2017]{paszke2017automatic}
Paszke, A., Gross, S., Chintala, S., Chanan, G., Yang, E., DeVito, Z., Lin, Z., Desmaison, A., Antiga, L., and Lerer, A. (2017).
\newblock Automatic differentiation in {PyTorch}.

\bibitem[Vaswani et~al., 2017]{vaswaniAttention2017a}
Vaswani, A., Shazeer, N., Parmar, N., Uszkoreit, J., Jones, L., Gomez, A.~N., Kaiser, {\L}., and Polosukhin, I. (2017).
\newblock Attention is all you need.
\newblock {\em Advances in neural information processing systems}, 30.

\bibitem[Wu et~al., 2020]{wu2020visual}
Wu, B., Xu, C., Dai, X., Wan, A., Zhang, P., Yan, Z., Tomizuka, M., Gonzalez, J., Keutzer, K., and Vajda, P. (2020).
\newblock {Visual Transformers: Token-based Image Representation and Processing for Computer Vision}.

\end{thebibliography}

}

%%%%%%%%%%%%%%%%%%%%%%%appendix%%%%%%%%%%%%%%%%%%%%%%%%%%%%%
% \clearpage
\section*{\uppercase{Appendix}}

\subsection{Ablation Study Results}\label{sec:abl_appendix}
As described in Section \ref{sec:results}, an ablation study was performed where the entire ViT model was trained alongside the final layers using the contrastive loss function. Comparing these results to those obtained with the fixed layers, we can see that the fully-trained model has a higher type I error rate and a lower $F_1$ score on both the cross-validation and held-out test sets. This indicates that the fully-trained model is overfitting the data to some extent, which is expected given the increased flexibility of the model. Overall, the results of the ablation study support our decision to use a fixed ViT backbone with contrastive learning. This approach appears to be more effective at learning a robust representation of the data and generalizing to new samples, as demonstrated by the superior performance of the fixed layers model on both the cross-validation and held-out test sets.

\begin{figure}[hbt!]
    \centering
    \begin{subfigure}{0.7\linewidth}
    \includegraphics[width=\linewidth]{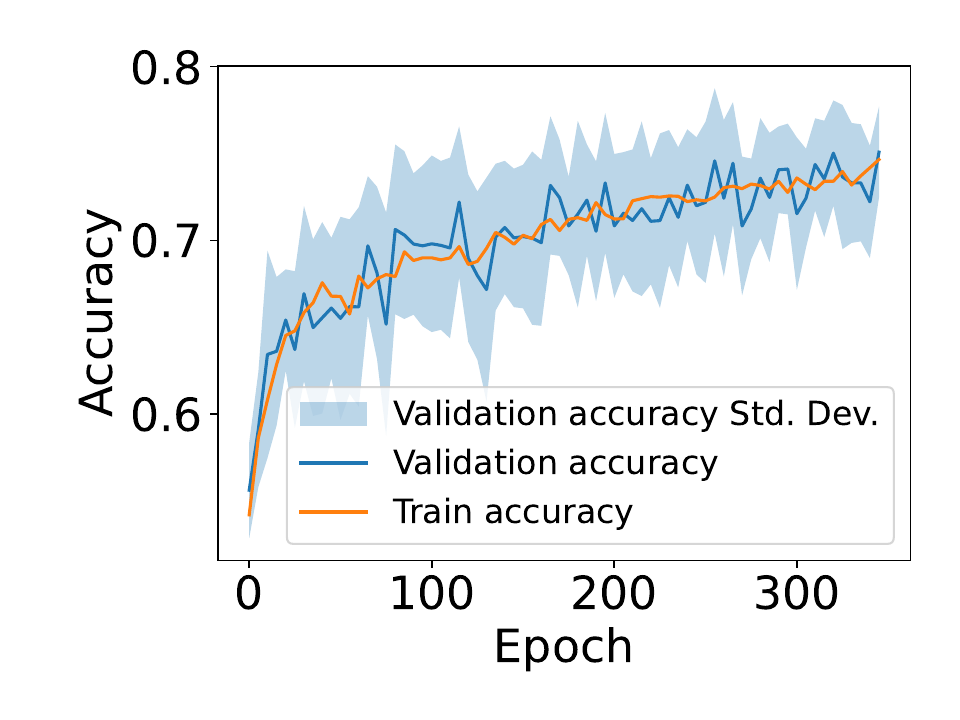}
    \caption{Mean train accuracy and validation accuracy.}
    \end{subfigure}
    \begin{subfigure}{0.7\linewidth}
    \includegraphics[width=\linewidth]{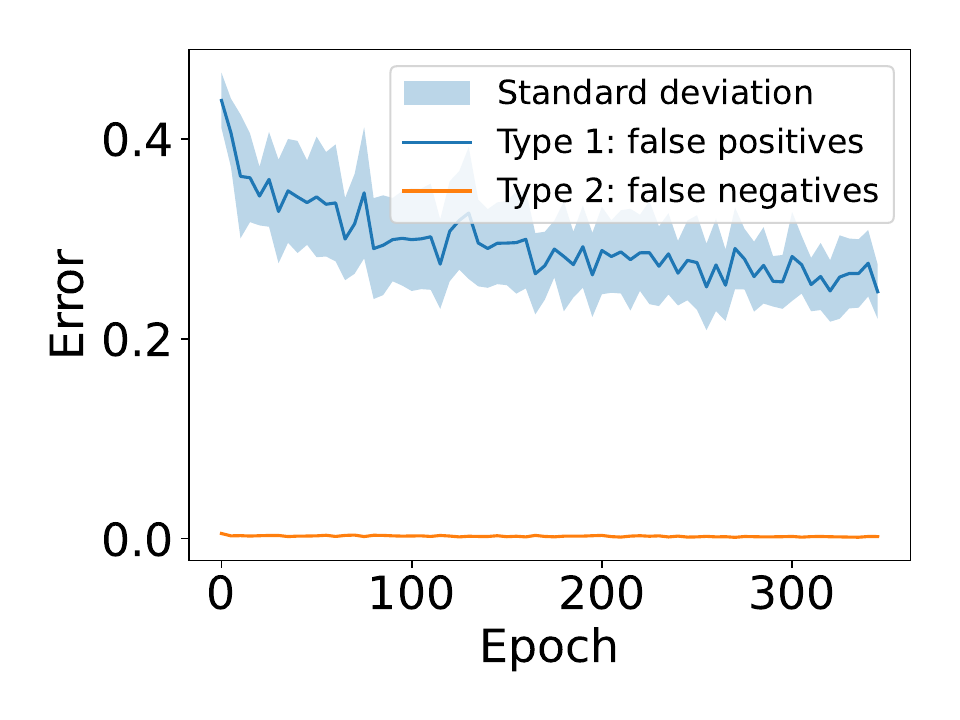}
    \caption{Type I and II errors of the model on the test set.}
    \end{subfigure}

    % \vspace{2mm}

    \begin{subfigure}{0.7\linewidth}
    \includegraphics[width=\linewidth]{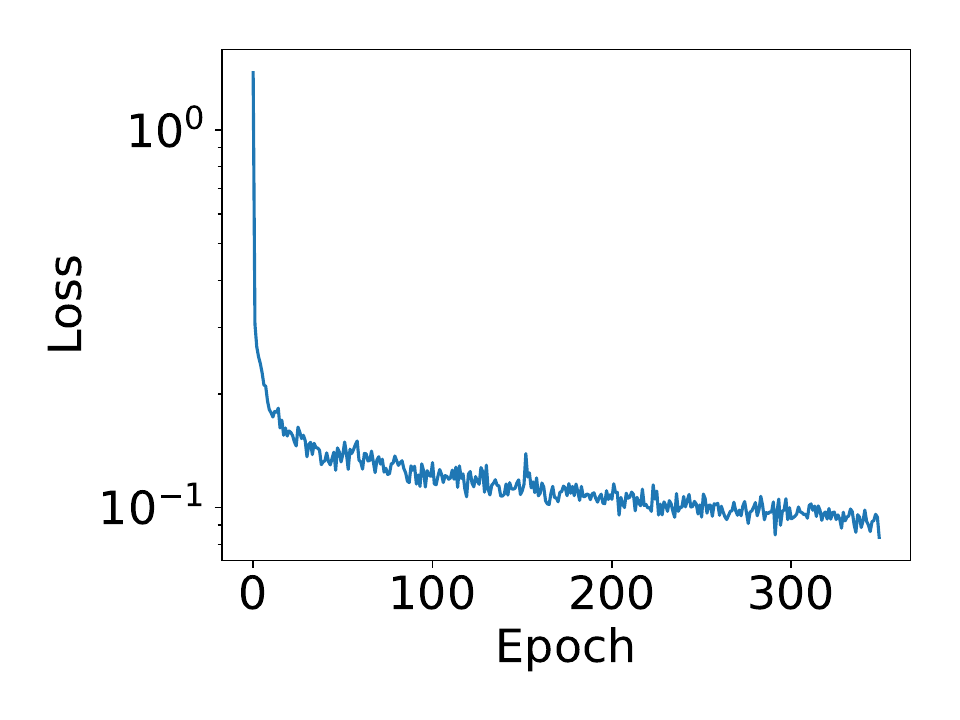}
    \caption{Loss of contrastive ViT Model.}
    \end{subfigure}
    \caption{Results of the ablation study averaged over three model runs. The data for accuracy was smoothed by averaging the values every five epochs. The data was collected and processed in the same manner as for the plots presented in Section~\ref{sec:results}.}
    \label{fig:abl_results}
\end{figure}

%%%%%%%%%%%%%%%%%%%%%%%DEMO%%%%%%%%%%%%%%%%%%%%%%%%%%%%%
\subsection{Model Demonstration} \label{sec:webapp_appendix}
To make the contrastive learning model available to a broader audience, we developed a web application that allows users to upload pictures of dogs and discover whether there are any similar dogs found in the system. The web application processes the image using the contrastive learning model and returns a list of pets along with their similarity score. 

\begin{figure}[hbt!]
    \centering
    \begin{subfigure}{0.6\linewidth}
    \includegraphics[width=\linewidth]{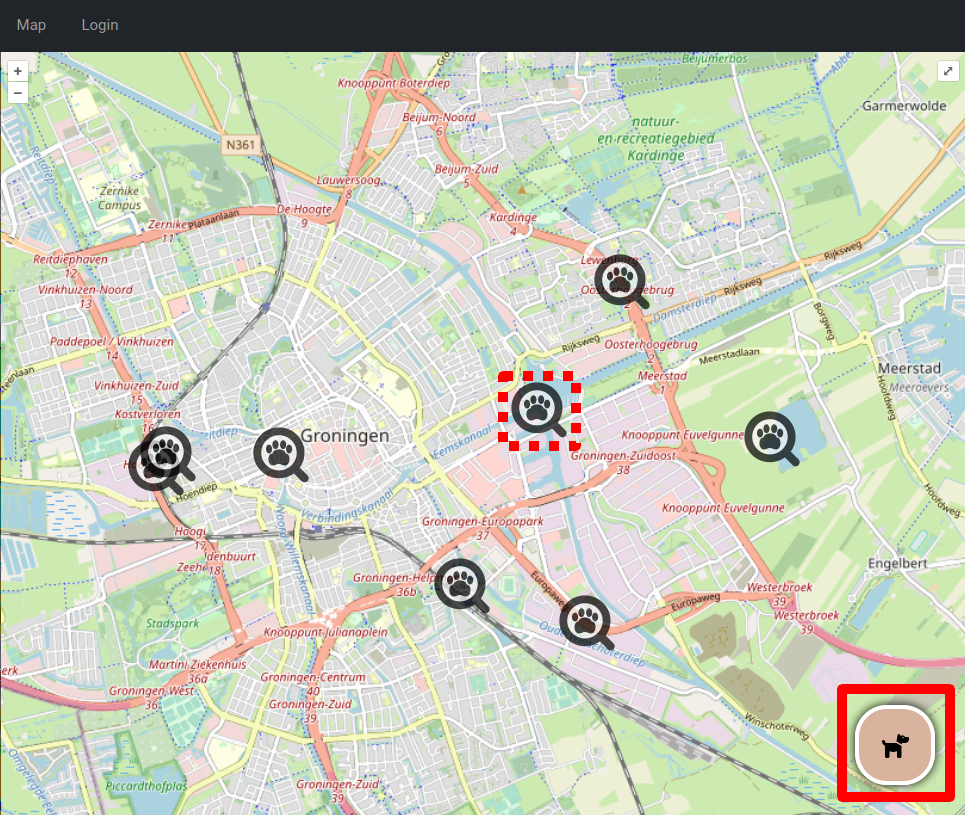}
    \caption{Map of pet sightings.}
    \vspace{2mm}
    \label{fig:webappMap}
    \end{subfigure}
    \begin{subfigure}{0.6\linewidth}
    \includegraphics[width=\linewidth]{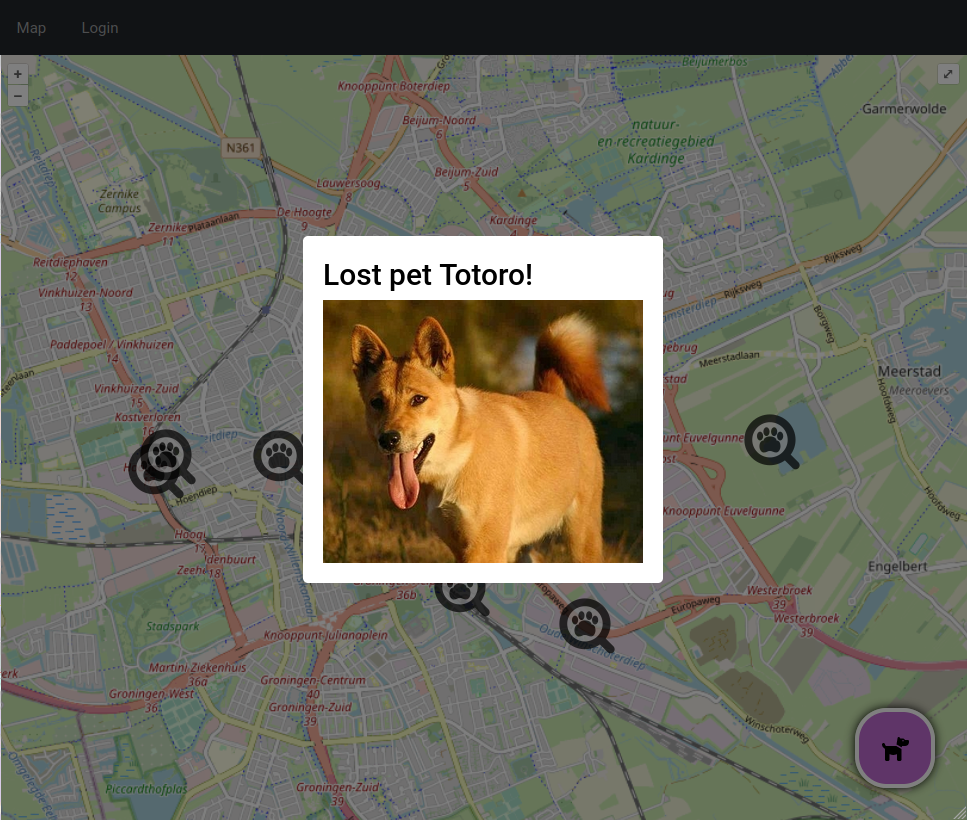}
    \caption{Image of sighted pet.}
    \vspace{2mm}
    \label{fig:webappLost}
    \end{subfigure}

    % \vspace{3mm}

    \begin{subfigure}{0.6\linewidth}
    \includegraphics[width=\linewidth]{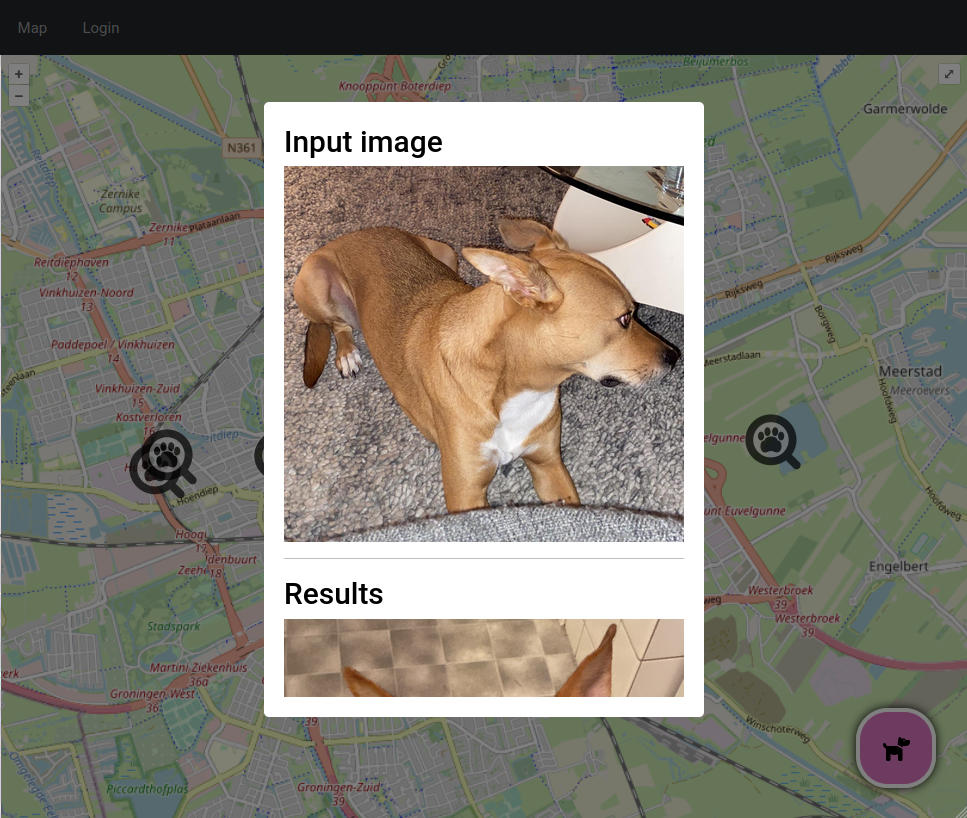}
    \caption{User uploaded a picture.}
    \vspace{2mm}
    \label{fig:webappResultInput}
    \end{subfigure}
    \begin{subfigure}{0.6\linewidth}
    \includegraphics[width=\linewidth]{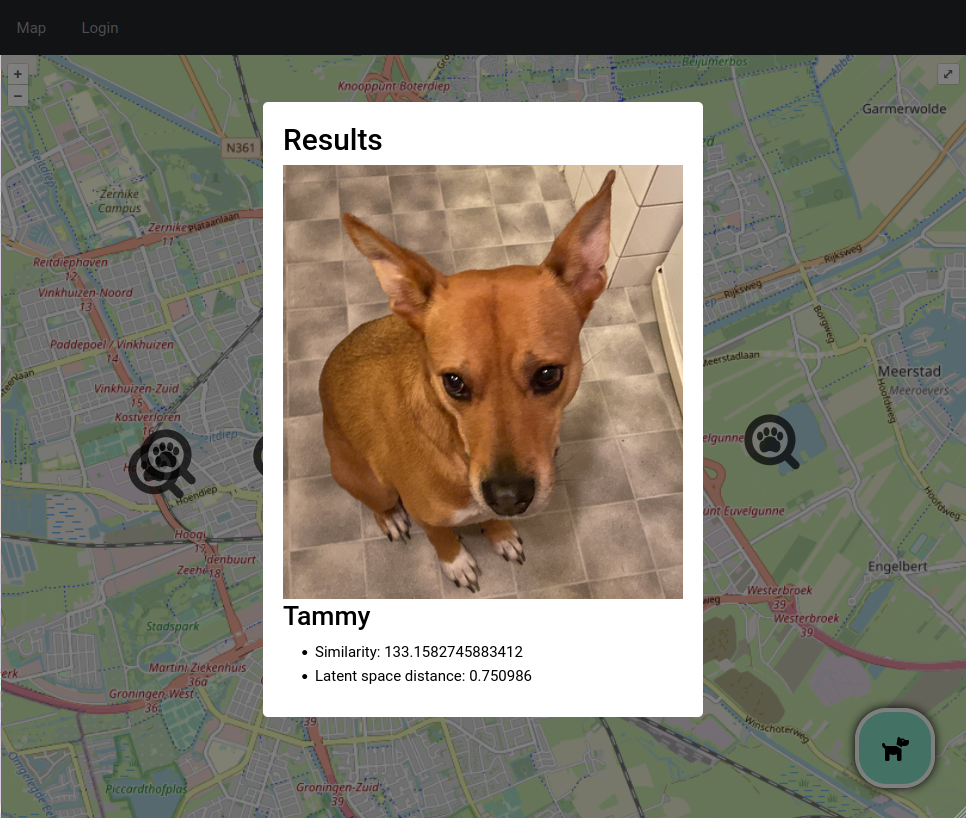}
    \caption{First hit of similar pets.}
    \vspace{2mm}
    \label{fig:webappResultOutput}
    \end{subfigure}
    \caption{Screenshots of the web application showing how users might interact with the website. Clicking the symbol highlighted by the red dashed outline (in \ref{fig:webappMap}) opens the pop-up shown in \ref{fig:webappLost}. Similarly, clicking the symbol highlighted by the solid red outline opens a picture upload dialog and displays the results as in \ref{fig:webappResultInput} and \ref{fig:webappResultOutput}. For code availability, please follow the following link: \href{https://github.com/vandrw/lostpaw-transformer}{https://github.com/vandrw/lostpaw-transformer}.}
    \label{fig:webapp}
\end{figure}

\end{document}